\documentclass[conference]{IEEEtran}
\IEEEoverridecommandlockouts
\usepackage{cite}
\usepackage[T1]{fontenc}
\usepackage{amsmath,amssymb,amsfonts}
\usepackage{algorithmic}
\usepackage{graphicx}
\usepackage{hyperref}
\usepackage{booktabs, makecell, tabularx, multirow, glossaries}
\usepackage{booktabs}
\usepackage[capitalize]{cleveref}
\usepackage{siunitx}
\usepackage{float}
\usepackage{tikz}
\usepackage[super]{nth}
\usepackage{pgfplots}
\pgfplotsset{compat=newest}
\usepgfplotslibrary{groupplots}
\usetikzlibrary{external}
\usepackage{array}
\usetikzlibrary{positioning}
\usepackage{url}
\usepackage{textcomp}
\usepackage{xcolor}
\def\BibTeX{{\rm B\kern-.05em{\sc i\kern-.025em b}\kern-.08em
    T\kern-.1667em\lower.7ex\hbox{E}\kern-.125emX}}

\usepackage{fancyhdr}

\begin{document}

\title{\LARGE \bf A Reinforcement Learning-Boosted Motion Planning Framework: Comprehensive Generalization Performance in Autonomous Driving}

\author{Rainer Trauth$^{1,*}$, Alexander Hobmeier$^{1}$ and Johannes Betz$^{2}$
\thanks{$^{1}$ The authors are with the Institute of Automotive Technology,
        Technical University of Munich, 85748 Garching, Germany; Munich Institute of Robotics and Machine Intelligence (MIRMI). They gratefully acknowledge the financial support from the Technical University of Munich - Bavarian Research Foundation (BFS)}
\thanks{$^{2}$The author is with the Professorship of Autonomous Vehicle Systems, Technical University of Munich, 85748 Garching, Germany; Munich Institute of Robotics and Machine Intelligence (MIRMI)}
\thanks{$^{*}$Corresponding first author: Rainer Trauth (rainer.trauth@tum.de)}
}

\maketitle

\begin{abstract}
This study introduces a novel approach to autonomous motion planning, informing an analytical algorithm with a reinforcement learning (RL) agent within a Frenet coordinate system. The combination directly addresses the challenges of adaptability and safety in autonomous driving. Motion planning algorithms are essential for navigating dynamic and complex scenarios. Traditional methods, however, lack the flexibility required for unpredictable environments, whereas machine learning techniques, particularly reinforcement learning (RL), offer adaptability but suffer from instability and a lack of explainability. Our unique solution synergizes the predictability and stability of traditional motion planning algorithms with the dynamic adaptability of RL, resulting in a system that efficiently manages complex situations and adapts to changing environmental conditions. Evaluation of our integrated approach shows a significant reduction in collisions, improved risk management, and improved goal success rates across multiple scenarios. The code used in this research is publicly available as open-source software and can be accessed at the following link: \url{https://github.com/TUM-AVS/Frenetix-RL}.


\end{abstract}

\begin{IEEEkeywords}
Adaptive algorithms, Autonomous vehicles, \\ Collision avoidance, Reinforcement learning, Robot learning
\end{IEEEkeywords}

\section{Introduction}
\label{sec:introduction}
While autonomous driving technology has raised considerable interest and enthusiasm, its real-world implementation has highlighted significant challenges as documented in various collision reports~\cite{coll_reports}. These challenges include navigating complex urban settings, managing unpredictable traffic and pedestrian behaviors, and making informed decisions in novel environments. 
Such unpredictability demands highly sophisticated and adaptable algorithms in the field of motion planning. Traditional analytical planning methods are often inadequate in handling the dynamic nature of real-world scenarios, emphasizing the critical need for enhanced decision-making capabilities and robust adaptability in autonomous driving systems to ensure safety and efficiency. In addition, analytical or rule-based models require extensive optimization through parameter tuning. This involves identifying and adjusting various settings and parameters fitted to specific scenarios. These adjustments are typically made through expert knowledge and numerical evaluation techniques. Notably, even minor parameter changes can noticeably impact the system's behavior. Adjusting this system can be both inefficient and costly. This becomes more evident when dealing with multiple configurations and variants.
Contemporary machine learning methods, especially RL, promise exceptional performance in complex scenarios. However, the effectiveness of the learning process is contingent upon the specific environment and training configurations used. Especially in autonomous driving, machine learning models for motion planning have low success rates or can only succeed in specific environments and scenarios like highway driving~\cite{Aradi2022,Zhou2022,Wang2021CR,Dinneweth2022,shalevshwartz2016safe}. Furthermore, long training times are required for complex scenarios, and problems can occur regarding Sim2Real~\cite{Aradi2022}. Moreover, the decision-making process of these agents often lacks inherent transparency, necessitating considerable effort in terms of validation and implementing safety measures to ensure reliability and trustworthiness in their actions~\cite{Minh2022,BARREDOARRIETA202082,Dinneweth2022}. Addressing these challenges is crucial, particularly in autonomous driving, where safety and reliability are paramount. 
\begin{figure}[t]
  \centering
  \includegraphics[width=0.475\textwidth]{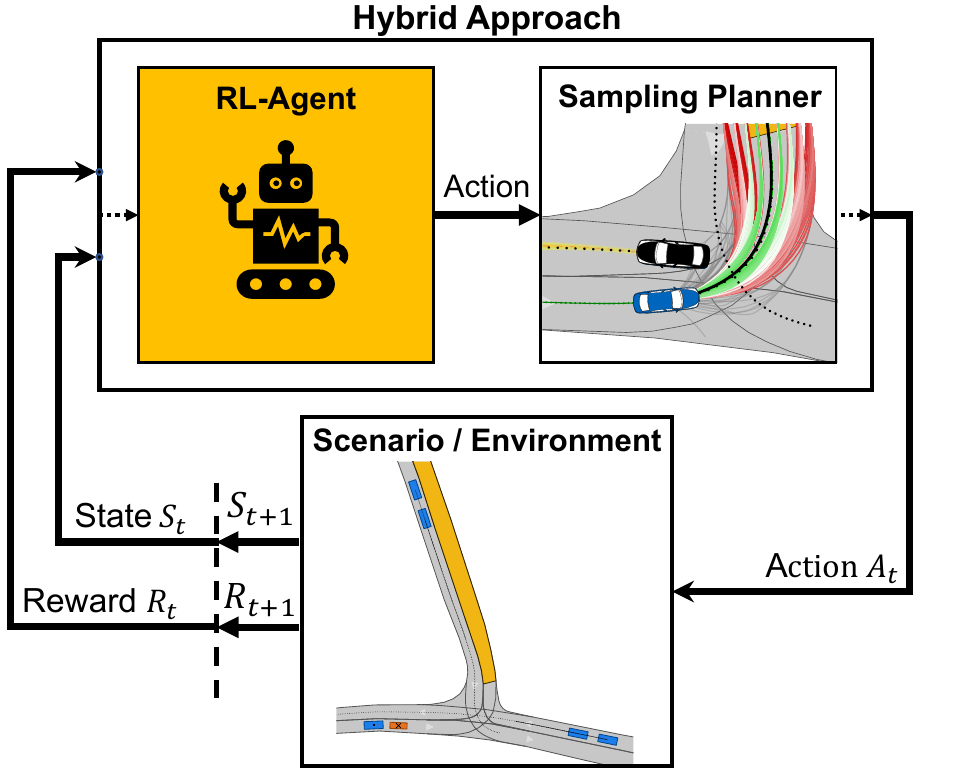}
  \caption{Hybrid reinforcement learning principle of a motion planning agent. The agent's action informs an analytical (e.g., sampling-based) method to achieve the goal.}
  \label{fig:fig1}
  \vspace{-0.5em}
\end{figure}
In contrast, hybrid methods combining analytical and machine learning models are expected to offer advantages in both areas. Therefore, we present a new approach to motion planning with a two-stage agent, shown in \cref{fig:fig1}.
In this methodology, the foundational robustness of analytic models is integrated with the dynamic learning capabilities of machine learning algorithms, enhancing both performance and adaptability in driving behavior contexts. This hybrid approach effectively bridges the gap between theoretical constructs and practical application, particularly in environments with complex, non-linear data patterns. Notably, these hybrid models often require less data for effective training, providing an advantage in data-poor scenarios~\cite{Zhou2022,Dong2023,Xiao2022,Xiao2022appl}. Furthermore, the efficient integration of safety methods and additional functionalities into the analytical planning algorithm is possible.
In summary, this work presents three main contributions:
\begin{enumerate}
	\item \textbf{A Hybrid motion planning methodology:} This encompasses the creation of a motion planning approach that integrates environment and prediction information within a Frenet coordinate system. The focus is on leveraging the strengths of the hybrid model to improve motion planning capabilities.
	\item \textbf{Performance analysis:} The new methodology will undergo extensive analysis to assess its efficiency, effectiveness, and improvement areas, providing valuable insights into the hybrid model approach's performance across different scenarios.
	\item \textbf{Open-source software:} An executable software framework will be available as open-source to integrate additional approaches.
\end{enumerate}

\section{Related work}
\label{sec:relatedwork}
Autonomous driving motion planning has been an area of intense research for many years. Several approaches are being developed to address the planning task of autonomous driving. Planning methods can be divided into the following categories at a high level~\cite{Teng2023,Dong2023,Zhou2022,Paden2016}:
Graph-based algorithms navigate through networks of nodes and edges to find structured paths~\cite{Teng2023,Zhou2022,Rowold2022}. 
Sampling-based methods explore a wide range of trajectories by generating numerous possibilities~\cite{werling.2010,werling.2012,Huang2023,Würsching2021}. 
Optimization-based planning methods aim to find the most effective trajectory by systematically evaluating various constraints and objectives, often using techniques like linear programming, dynamic programming, or gradient-based optimization~\cite{Teng2023,Zhou2022,Cao2023,Li2023,Jeong2023,Zhu2023}. 
Furthermore, algorithms that utilize artificial intelligence are developed to provide high adaptability in dynamic environments, demonstrating the integration of adaptive computational techniques in this domain~\cite{Teng2023,Zhou2022,Dong2023,Elallid2022,shalevshwartz2016safe}.
There are several machine learning models in the literature that learn to control the steering wheel and acceleration. These models are almost exclusively trained using specific scenarios such as highway driving or decision-making agents~\cite{Wang2021CR,Mirchevska2022,Zhang2021,Al-Sharman2023,Gu2023,Klimke2023}. Although the models show improvement, the success rate for more difficult scenarios is too low, especially for real-world application~\cite{Wang2021CR}. 
Learning human-like behavior is also investigated through inverse RL~\cite{Huang2023-IRL,Trauth-IRL}. The driving behavior of certain characteristics can be learned and adopted. However, this does not lead to a fundamentally higher success rate. 
The contribution of Xiao et al.~\cite{Xiao2022appl,Xiao2022} explores how iterative learning and human feedback can improve navigation in complex environments for autonomous robots. By integrating these elements into traditional navigation systems, the study demonstrates potential performance improvements while keeping the systems safe and interpretable. This research offers a notable perspective on developing adaptive navigation systems in robotics. The results, while promising, primarily serve as a proof of concept. They do not incorporate complex public road environments or account for the prediction uncertainties of other road users. Moreover, the approach does not integrate a complex analytical planning algorithm; instead, it relies on machine learning to assimilate parameter settings based on expert knowledge.
Yu et al.~\cite{Yu2023} present a framework combining RL with a rapidly exploring random tree for autonomous vehicle motion planning. It focuses on efficiently controlling vehicle speed and ensuring safety, using deep learning techniques to adapt to varying traffic conditions. The primary issue with the approach is its slow convergence rate in high-dimensional state spaces, which compromises its real-time applicability. Furthermore, the method is designed for only certain scenarios, limiting its generalizability.
Other research employs RL to determine optimal switching points for executing movements through an analytical model. This approach is applied in scenarios such as timing lane changes and facilitating interaction behaviors among different road users~\cite{Albarella2023,Jafari2023,Klimke2023}.
The current scientific landscape shows a gap in exploring a hybrid approach that combines machine learning with a powerful analytical algorithm for trajectory planning that ensures high success rates, real-time capability, interpretability, and the integration of additional safety features. Strengths and weaknesses could be investigated with such a concept independent of supervised learning datasets.

\section{Methodology}
\label{sec:methodology}
This section presents the combination of the analytical sampling-based trajectory planner architecture and the RL design for developing the hybrid motion planning method. 
\subsection{Sampling-based Motion Planner}
\label{sec:frenetix-planner}
\begin{figure}[ht!]
  \centering
  \hspace{1em}
  \includegraphics[width=0.35\textwidth]{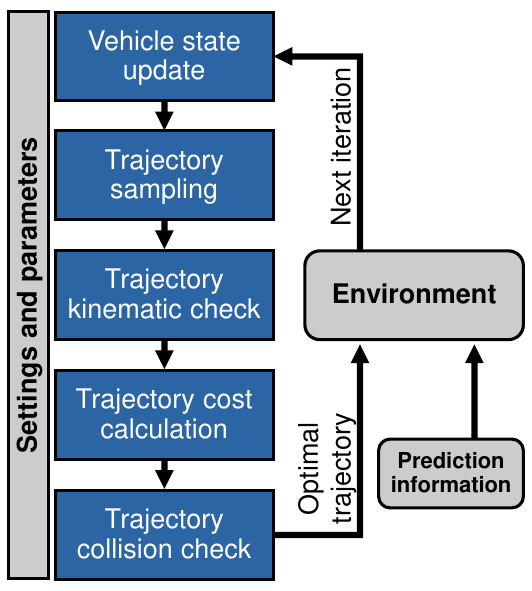}
  \setlength{\abovecaptionskip}{-4pt}
  \caption{Frenetix Motion Planner: sampling-based motion planning procedure for one timestep including prediction information and environment state update.}
  \label{fig:frenetix_planner}
  \vspace{-0.8em}
\end{figure}

The analytical trajectory planning algorithm used is a sampling-based approach in the Frenet coordinate system according to the concept of Werling et al.~\cite{werling.2010,werling.2012}. The software is online and available on GitHub\footnote{\url{https://github.com/TUM-AVS/Frenetix-Motion-Planner}}\footnote{\url{https://github.com/TUM-AVS/Frenetix}}. We use a neural network-based algorithm to predict other vehicles in the scenario~\cite{walenet}. The procedure of the algorithm during one timestep is shown in \cref{fig:frenetix_planner}.
The procedure can be summarised into the following main stages: 
\begin{itemize}
    \item \textbf{Vehicle state update:} The vehicle updates all states concerning the Frenet coordinate system using the ego, prediction, and environment information.
    \item \textbf{Trajectory sampling:} The algorithm samples possible trajectories regarding a sampling matrix. We use time-, velocity-, and lateral-sampling metrics to create different trajectory schemes dependent on the current ego vehicle state~\cite{werling.2010,werling.2012}.
    \item \textbf{Trajectory kinematic check:} The generated trajectories are checked for kinematic feasibility as a function of a single-track model and the vehicle parameters.
    \item \textbf{Trajectory cost calculation:} We use different cost metrics like collision probability, jerk, distance to reference path, and velocity offset costs to differentiate between the performance of different trajectories. We integrate the collision probability cost with other obstacles from the prediction information~\cite{walenet}. The trajectory generation is implemented in C++\footnote{\url{https://github.com/TUM-AVS/Frenetix}} to reduce calculation time and accelerate the training process.
    \item \textbf{Trajectory collision check:} The lowest cost trajectory is analyzed for possible collisions with the lane boundary and other obstacles~\cite{DrivabilityChecker}. This step takes place after the cost calculation step for computational efficiency. The first collision-free trajectory sorted by absolute cost is the optimal trajectory to update the current vehicle state.
\end{itemize}
The vehicle's status is updated based on the optimal trajectory calculated for each successive timestep. The trajectories encompass a horizon of \SI{3}{\second}. The timestep discretization of the simulation \SI{0.1}{\second}.

\subsection{Reinforcement Learning Procedure}
\label{sec:rl-method}
In this section, we are integrating an RL algorithm, which optimizes the trajectory selection process for the presented sampling-based trajectory planner of \cref{sec:frenetix-planner}. For the customized environment and training process, we use gymnasium\footnote{\url{https://github.com/Farama-Foundation/Gymnasium}} and stable-baselines3\footnote{\url{https://github.com/DLR-RM/stable-baselines3}}. For the agent's simulation environment, we utilize CommonRoad~\cite{commonroad,Wang2021CR}. 
The optimization is performed by Proximal Policy Optimization (PPO)~\cite{PPO}, an RL algorithm that balances exploration and exploitation by clipping the policy update. It avoids large policy updates that could collapse performance, making training more stable and reliable. The core of the PPO algorithm is encapsulated in \cref{eq:ppo}~\cite{PPO}:
\begin{align}
    L^{CLIP}(\theta) &= \hat{\mathbb{E}}_t \left[ \min(r_t(\theta) \hat{A}_t, \right. \nonumber \\
    &\quad \left. \text{clip}(r_t(\theta), 1 - \epsilon, 1 + \epsilon) \hat{A}_t) \right]
\label{eq:ppo}
\end{align}

This equation represents the clipped surrogate objective function, which is crucial for the efficiency and stability of the PPO algorithm. Here, \( \theta \) represents the policy parameters, \( \hat{\mathbb{E}}_t \) is the empirical expectation over timesteps, \( r_t(\theta) \) signifies the probability ratio under new versus old policies, \( \hat{A}_t \) denotes the estimated advantage at time \( t \), and \( \epsilon \) is a critical hyperparameter controlling the clipping in the objective function. 
We use Recurrent PPO Optimization with MlpLstmPolicy to process temporal relationships and information. The conventional PPO architecture is extended with Long Short-Term Memory (LSTM) networks~\cite{Hochreiter1997}, a recurrent neural network suited for dynamic, time-series data. This approach is effective in sequential data and partially observable environments. The LSTM component can be formulated as follows:
\begin{itemize}
    \item \textbf{LSTM state update}: At each timestep $t$, the LSTM updates its hidden state $h_{t-1}$ and cell state $c_t$ based on the current input $x_t$, previous hidden state $h_{t-1}$, and previous cell state $c_{t-1}$. Expressed as: $(h_t,c_t)=LSTM(x_t,h_{t-1},c_{t-1})$
    \item \textbf{Policy and value function}: The updated hidden state $h_t$ is then utilized by the policy network $\pi(a_t|s_t,h_t)$ and the value network $V(s_t, h_t)$, where $a_t$ is the action and $s_t$ is the state at time $t$. This integration enables the network to remember past states, enhancing decision-making in complex environments.
\end{itemize}

\begin{figure*}[t]
  \centering
  \setlength{\abovecaptionskip}{-4pt}
  \includegraphics[width=0.8\textwidth]{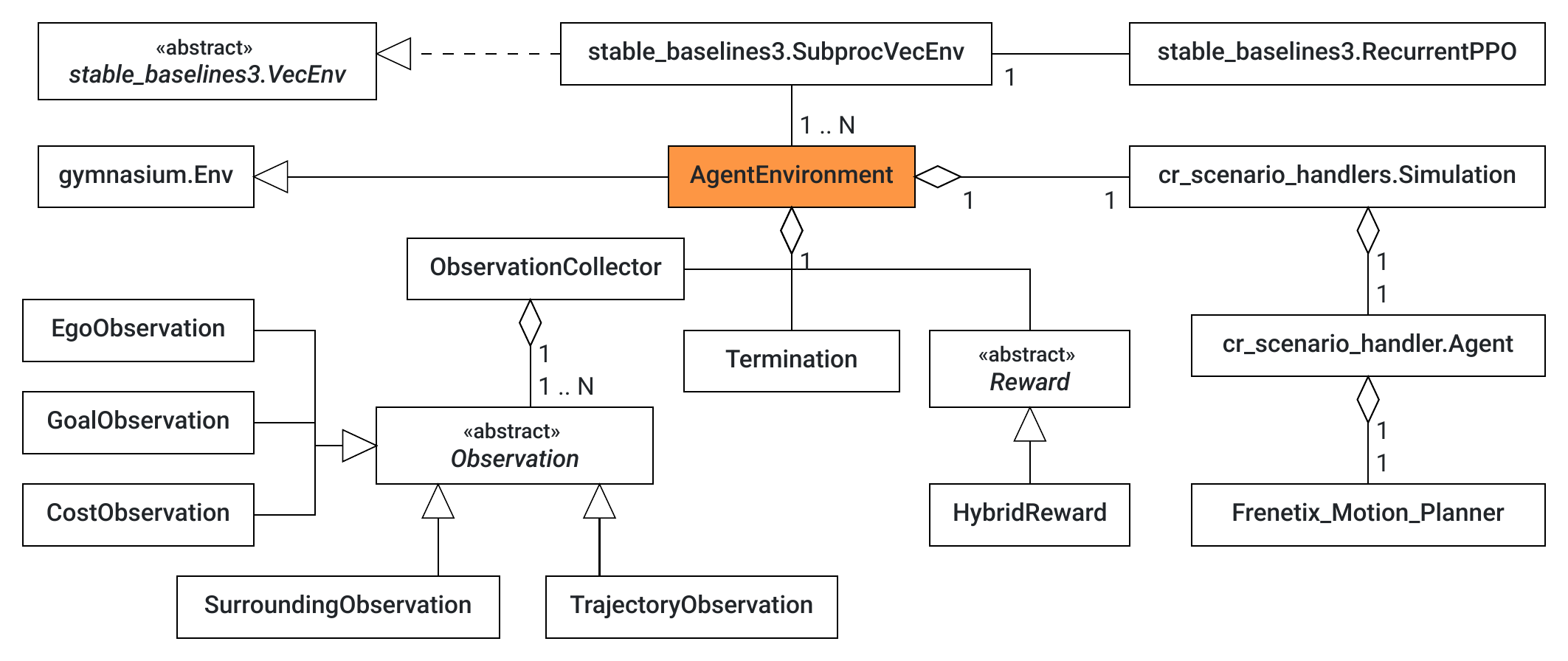}
  \caption{Class diagram of the learning process structure.}
  \label{fig:classdiagram}
\end{figure*}

In order to initiate the optimization process, it is necessary to first design several key components: the observational space, the criteria for termination, the structure of the reward system, and the definition of the agent's action space. \cref{fig:classdiagram} shows a class diagram that provides an overview of the functions integral to the training procedure.

\textbf{Observation space:}
The observation space is divided into the categories and observations in \cref{tab:observations}.
\begin{table}[ht]
    \caption{Observation space of the learning process.}
    \centering
    \renewcommand{\arraystretch}{1.5}
    \begin{tabular}{m{1.4cm} p{5.9cm}} 
    \hline
    \textbf{Categories}              & \textbf{Observations} \\ \hline\hline
    
    Ego  & Velocity, acceleration, jerk, steering, orientation, yaw, distance to reference path   \\ 
    
    Goal & Distance to goal, remaining time, goal reached, timeout, target velocity    \\ 
    
    Surrounding      & Adjacent lanes, the direction of adjacent lanes, obstacle information     \\ 
    
    Trajectory       & Percentage of feasible trajectories, validity of trajectories, ego risk, obstacle (\nth{3}-party) risk    \\ 
    
    Cost    & Cost of optimal trajectory, cost mean \& variance of all trajectories, cost collision probability    \\ 
    \hline
    \end{tabular}
    \label{tab:observations}
\end{table}

\begin{figure}[ht]
  \centering
  \includegraphics[width=0.4\textwidth]{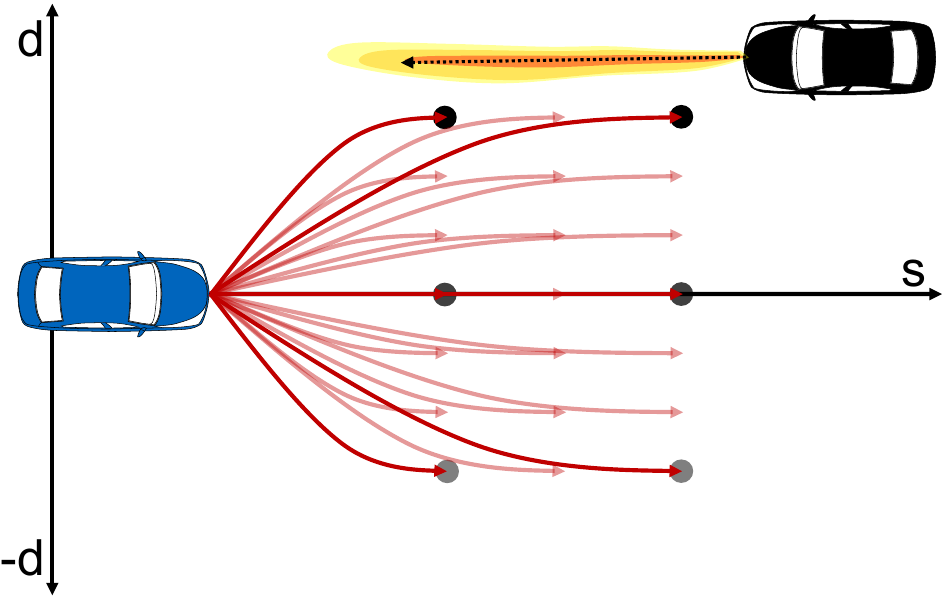}
  \caption{Trajectory cost observation space. The cost calculations of the outermost trajectories perceive additional cost information at each timestep.}
  \label{fig:trajectory_perception}
\end{figure}

The categories can be segmented into various types: those originating from the ego vehicle, those pertinent to achieving the goal region, surrounding information, trajectory details, and cost information related to the sampled trajectories. 
Unlike other systems that merely assume direct vehicle control~\cite{Wang2021CR}, our methodology provides supplementary data that enhances the observation space. The several hundred sampled trajectories of the trajectory planning algorithm contain additional information through the calculation steps in \cref{fig:frenetix_planner}. Key elements of this data include the count of kinematically feasible trajectories, the associated risk level of each trajectory, and their respective cost distributions. Furthermore, we use our concept, depicted in \cref{fig:trajectory_perception}, to address collision probability perception.

The schematic illustration presents the sampled trajectories. We can construct a grid by employing time-, speed-, and lateral-dependent sampling. This grid allows us to analyze the variation in collision probability costs associated with the outermost trajectories, enriching the observation space. Such an approach enables the mapping of differences and correlations over time. In the illustration, trajectories in the positive lateral d-direction are attributed to higher collision probability costs than those in the negative d-direction.

\textbf{Action space:}
\cref{fig:fig1} shows the connection between the analytical trajectory planner and the RL agent. The agent learns the action, which is the cost weight of the trajectory planner. Theoretically, any adjustment can be passed to the trajectory planner. In our case, we investigate the adjustment of cost weights to prove the concept. To achieve a harmonic behavior, the agent can reduce or increase the current cost weights of the trajectory planner. \cref{eq:cost_weights} shows the action space of the agent regarding each cost term $i$ in timestep~$t$.
\begin{equation}
    \omega_{i}^{t} = \omega_{i}^{\text{min}} \leq  \omega_{i}^{prev} +  \omega_{i}^{action} \leq \omega_{i}^{\text{max}}
    \label{eq:cost_weights}
\end{equation}

Consider $\omega_{i}^{\text{action}}$, a floating-point value in the range $[a_{\text{min}}, a_{\text{max}}]$. Here, $\omega_{i}^{\text{min}}$ and $\omega_{i}^{\text{max}}$ represent the predefined minimum and maximum values of the absolute cost term, respectively. Additionally, $\omega_{i}^{prev}$ denotes the weight from the previous timestep, while $\omega_{i}^{action}$ signifies the current action of the algorithm. It is important to note that after each execution, the cost terms are reset to their default values.

\textbf{Reward design:}
The training process requires the reward configuration and is crucial for success and driving behavior.
The rewards we use for the learning process are shown in \cref{tab:rewards}:
\begin{table}[ht]
    \caption{Hybrid Reward of the PPO Training Process.}
    \centering
    \renewcommand{\arraystretch}{1.2}
    \begin{tabular}{c c}
    \hline
    \textbf{Sparse reward} & \textbf{Dense reward} \\ \hline\hline
    Goal reached   & Distance to reference path   \\ 
    Goal reached (faster than target time)   & Difference to target velocity    \\ 
    Goal reached (slower than target time)   & S-distance to goal   \\ 
    Collision   & Standard cost difference    \\ 
    No feasible solution   & Ego risk   \\ 
    Scenario timeout   & Obstacle risk  \\
    \hline
    \end{tabular}
    \label{tab:rewards}
\end{table}
Our approach uses a hybrid reward system to enhance training efficiency: termination reward and sparse reward. The termination reward is crucial for finishing scenarios successfully, while a sparse reward guides vehicle behavior. The main goal is to minimize collisions, especially influenced by the termination reward. In addition, sparse rewards are required to optimize driving performance and behavior, such as satisfying comfort metrics or minimizing overall driving risk. 
The vehicle can finalize a scenario in six distinct ways. Each scenario has a distinct time horizon, establishing a window within which the goal can be achieved. This allows for reaching the goal either more quickly or slowly than the allotted time interval, depending on the vehicle's performance~\cite{commonroad}. Scenarios may end due to a collision with obstacles or road boundaries or if the vehicle fails to find a valid trajectory at any timestep. Additionally, if the vehicle stops without reaching the goal, the scenario will automatically terminate after exceeding a specific time limit. The optimal performance involves closely adhering to the reference path, maintaining the designated speed, maximizing progress towards the target distance, and minimizing risk~\cite{geisslingerconcept,GeißlerMaxRisk}. We are integrating a cost regulation term to enhance stability in the vehicle's actions. This addition aims to prevent excessive fluctuations in actions, promoting smoother and more harmonious driving behavior. We use the absolute difference between the current action and the default cost settings of the trajectory planner.

\section{Results \& Analysis}
\label{sec:results}
This section shows the model's training, selected testing scenarios, and results. We will explore the model qualitatively and quantitatively, highlighting the practicality of the concept. We will investigate the differences between the standalone default analytical trajectory planner (DP) and the proposed hybrid planner (HP).

\subsection{Environment and Training Setup}
We use T-junction scenarios (see \cref{fig:pathdeviation}) for the training process as they exhibit complex and critical interaction dynamics with other vehicles~\cite{commonroad}. Various scenarios in the data set offer a certain degree of variability to reduce the risk of overfitting. For the training and execution of the model, the computational resources include an AMD 7950x processor, an NVIDIA GeForce RTX 4090 graphics card, and 128GB of RAM. The hyperparameters used in our study are shown in \cref{tab:hyperparameters}.
\begin{table}[ht]
\centering
\caption{Hyperparameters of the PPO Algorithm.}
\renewcommand{\arraystretch}{1.2}
\begin{tabular}{l c}
    \hline
    \textbf{Hyperparameter} & \textbf{Value} \\ \hline\hline
    Learning Rate           & \SI{0.0003}{} \\ 
    Clipping parameter \(\epsilon\) & \SI{0.1}{} \\ 
    Discount factor \(\gamma\) & \SI{0.99}{} \\ 
    GAE \(\lambda\)             & \SI{0.97}{} \\
    Batch Size              & \SI{2352}{} \\ 
    Epochs                  & \SI{5}{} \\ 
    Entropy coefficient     & \SI{0.01}{} \\ \hline
\end{tabular}
\label{tab:hyperparameters}
\end{table}

The training is parallelized to the number of cores and requires approximately \SI{24}{\hour} for \SI{7}{} million timesteps. The data is classified into training set (\SI{75}{\percent}), validation set (\SI{15}{\percent}), and test set (\SI{10}{\percent}). The best model is selected based on the reward function in a series of evaluation scenarios. The training converges after \SI{2}{}-\SI{3}{} million training steps, depending on the settings. We use hyperparameter tuning because the training results are highly dependent on it.

\subsection{Risk-aware Trajectory Planning}
First, we look at the risk behavior of the learned agent, for which we also set a reward (see \cref{tab:rewards}) to optimize the agent's behavior. In addition to the success rate, the risk in autonomous driving is also decisive in evaluating the safety of the algorithm. The risk is subsequently calculated by multiplying the maximum collision probability $p$ of a trajectory $\mathcal{T}$ by the harm $H$ incurred~\cite{geisslingerconcept,geisslingernature,GeißlerMaxRisk}.
\begin{equation}
\begin{split}
	R(\mathcal{T}) = \mathrm{max}(p(\mathcal{T})H(\mathcal{T}))
\end{split}
\label{eq:trajrisk}
\end{equation} 

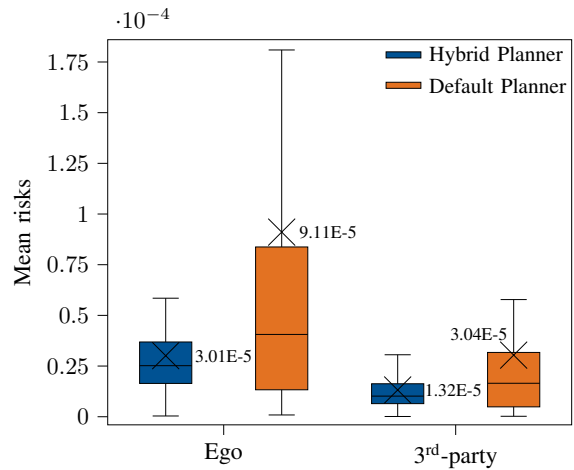
\begin{figure}[ht]
  \centering
  \scalebox{0.9}{
\begin{tikzpicture}

\definecolor{chocolate22711434}{RGB}{227,114,34}
\definecolor{darkgray176}{RGB}{176,176,176}
\definecolor{black}{RGB}{0,0,0}
\definecolor{teal082147}{RGB}{0,82,147}

\begin{axis}[
legend cell align={left},
legend columns=3,
legend style={
  fill opacity=0.8,
  draw opacity=1,
  text opacity=1,
  at={(1,1.22)},
  anchor=south west,
  draw=none
},
tick align=outside,
tick pos=left,
x grid style={darkgray176},
xmin=0.5, xmax=4.5,
xtick style={color=black},
xtick={1.5, 3.5}, 
xticklabels={Ego, \nth{3}-party}, 
y grid style={darkgray176},
ylabel={Mean risks},
ymin=-4.e-06, ymax=0.000186,
ytick={0.0,0.000025,0.00005,0.000075,0.0001,0.000125,0.00015,0.000175}, 
ytick style={color=black}
]
\path [draw=black, fill=teal082147]
(axis cs:0.775,1.64284873950871e-05)
--(axis cs:1.225,1.64284873950871e-05)
--(axis cs:1.225,3.68755704123054e-05)
--(axis cs:0.775,3.68755704123054e-05)
--(axis cs:0.775,1.64284873950871e-05)
--cycle;
\addplot [black, forget plot]
table {%
1 1.64284873950871e-05
1 3.7199077758795e-07
};
\addplot [black, forget plot]
table {%
1 3.68755704123054e-05
1 5.84700742847146e-05
};
\addplot [black, forget plot]
table {%
0.8875 3.7199077758795e-07
1.1125 3.7199077758795e-07
};
\addplot [black, forget plot]
table {%
0.8875 5.84700742847146e-05
1.1125 5.84700742847146e-05
};
\path [draw=black, fill=chocolate22711434]
(axis cs:1.775,1.32938709481537e-05)
--(axis cs:2.225,1.32938709481537e-05)
--(axis cs:2.225,8.37511627802008e-05)
--(axis cs:1.775,8.37511627802008e-05)
--(axis cs:1.775,1.32938709481537e-05)
--cycle;
\addplot [black, forget plot]
table {%
2 1.32938709481537e-05
2 8.76736039593867e-07
};
\addplot [black, forget plot]
table {%
2 8.37511627802008e-05
2 0.0001809306647252
};
\addplot [black, forget plot]
table {%
1.8875 8.76736039593867e-07
2.1125 8.76736039593867e-07
};
\addplot [black, forget plot]
table {%
1.8875 0.0001809306647252
2.1125 0.0001809306647252
};
\path [draw=black, fill=teal082147]
(axis cs:2.775,6.46047793671627e-06)
--(axis cs:3.225,6.46047793671627e-06)
--(axis cs:3.225,1.62644546392576e-05)
--(axis cs:2.775,1.62644546392576e-05)
--(axis cs:2.775,6.46047793671627e-06)
--cycle;
\addplot [black, forget plot]
table {%
3 6.46047793671627e-06
3 1.23315209433744e-07
};
\addplot [black, forget plot]
table {%
3 1.62644546392576e-05
3 3.05729423516334e-05
};
\addplot [black, forget plot]
table {%
2.8875 1.23315209433744e-07
3.1125 1.23315209433744e-07
};
\addplot [black, forget plot]
table {%
2.8875 3.05729423516334e-05
3.1125 3.05729423516334e-05
};
\path [draw=black, fill=chocolate22711434]
(axis cs:3.775,4.87348840634173e-06)
--(axis cs:4.225,4.87348840634173e-06)
--(axis cs:4.225,3.17186888393194e-05)
--(axis cs:3.775,3.17186888393194e-05)
--(axis cs:3.775,4.87348840634173e-06)
--cycle;
\addplot [black, forget plot]
table {%
4 4.87348840634173e-06
4 2.22445156756201e-07
};
\addplot [black, forget plot]
table {%
4 3.17186888393194e-05
4 5.77963863569876e-05
};
\addplot [black, forget plot]
table {%
3.8875 2.22445156756201e-07
4.1125 2.22445156756201e-07
};
\addplot [black, forget plot]
table {%
3.8875 5.77963863569876e-05
4.1125 5.77963863569876e-05
};
\addplot [black, forget plot]
table {%
0.775 2.51847634958444e-05
1.225 2.51847634958444e-05
};
\addplot [black, mark=x, mark size=8, mark options={solid,fill=black}, only marks, forget plot]
table {%
1 3.01079635966026e-05
};
\addplot [black, forget plot]
table {%
1.775 4.05995155335696e-05
2.225 4.05995155335696e-05
};
\addplot [black, mark=x, mark size=8, mark options={solid,fill=black}, only marks, forget plot]
table {%
2 9.10846035814713e-05
};
\addplot [black, forget plot]
table {%
2.775 1.0172931512908e-05
3.225 1.0172931512908e-05
};
\addplot [black, mark=x, mark size=8, mark options={solid,fill=black}, only marks, forget plot]
table {%
3 1.32001303064753e-05
};
\addplot [black, forget plot]
table {%
3.775 1.65074325745492e-05
4.225 1.65074325745492e-05
};
\addplot [black, mark=x, mark size=8, mark options={solid,fill=black}, only marks, forget plot]
table {%
4 3.0397225316135e-05
};
\draw (axis cs:1.2,2.7e-05) node[
  scale=0.75,
  anchor=base west,
  text=black,
  rotate=0.0
]{3.01E-5};
\draw (axis cs:2.1,8.8e-05) node[
  scale=0.75,
  anchor=base west,
  text=black,
  rotate=0.0
]{9.11E-5};
\draw (axis cs:3.18,1e-05) node[
  scale=0.75,
  anchor=base west,
  text=black,
  rotate=0.0
]{1.32E-5};
\draw (axis cs:3.4,3.8e-05) node[
  scale=0.75,
  anchor=base west,
  text=black,
  rotate=0.0
]{3.04E-5};

\draw[fill=teal082147] (axis cs:2.9,0.000177) rectangle (3.2,0.000180);
\node[align=left, anchor=west] at (axis cs:3.2,0.000178) {\small Hybrid Planner};

\draw[fill=chocolate22711434] (axis cs:2.9,0.000162) rectangle (3.2,0.000165);
\node[align=left, anchor=west] at (axis cs:3.2,0.000163) {\small Default Planner};

\end{axis}

\end{tikzpicture}
  }
  \caption{Mean ego and \nth{3}-party risks over various scenarios according to~\cite{geisslingerconcept}. Blue indicates the HP, and orange the DP.}
  \label{fig:mean_top_ego_risk}
\end{figure}

Our evaluation encompasses \SI{64}{} distinct scenarios to assess risk levels. We gain valuable insights into the overall safety landscape by calculating the mean risk across all scenarios. Notably, the results indicate a reduced risk for the ego vehicle and \nth{3}-party road users, highlighting improved road safety. 
\cref{fig:mean_top_ego_risk} shows the ego-vehicle risk of the scenarios and the \nth{3}-party risks. Blue indicates the HP and orange is the DP. The HP shows only about \SI{33}{\percent} of the risk for the ego vehicle compared to the DP.

The agent's reward for reducing the risk has a sustainable influence on the selection process of the trajectories. Our analysis demonstrates that, despite the complexity of numerous target variables, the vehicle can modify its behavior. It is crucial to emphasize the importance of carefully selecting reward terms within this framework. An overly aggressive pursuit of risk reduction through reward mechanisms may lead to scenarios where the vehicle opts to halt completely in certain situations. To mitigate this, we have incorporated a specific reward term, as delineated in \cref{tab:rewards}, which ensures adherence to the designated target speed, thereby balancing safety with operational efficiency in a controlled manner.
The risk is calculated based on the selected trajectory and depends on the planning horizon. The DP accepts a significantly higher risk for a short period and only reacts to the reduction once the risk has been recognized. On the other hand, the model presented here recognizes risky situations before they occur through environmental and obstacle information. The risk is significantly reduced both in absolute terms and in terms of its duration. By slowing down in advance, it can also be determined that the risk peaks occur with a time delay to the risk peaks of the DP.

\subsection{Adaption of the Agent's Driving Behavior}
\begin{figure}[ht]
  \centering
  \includegraphics[width=0.475\textwidth]{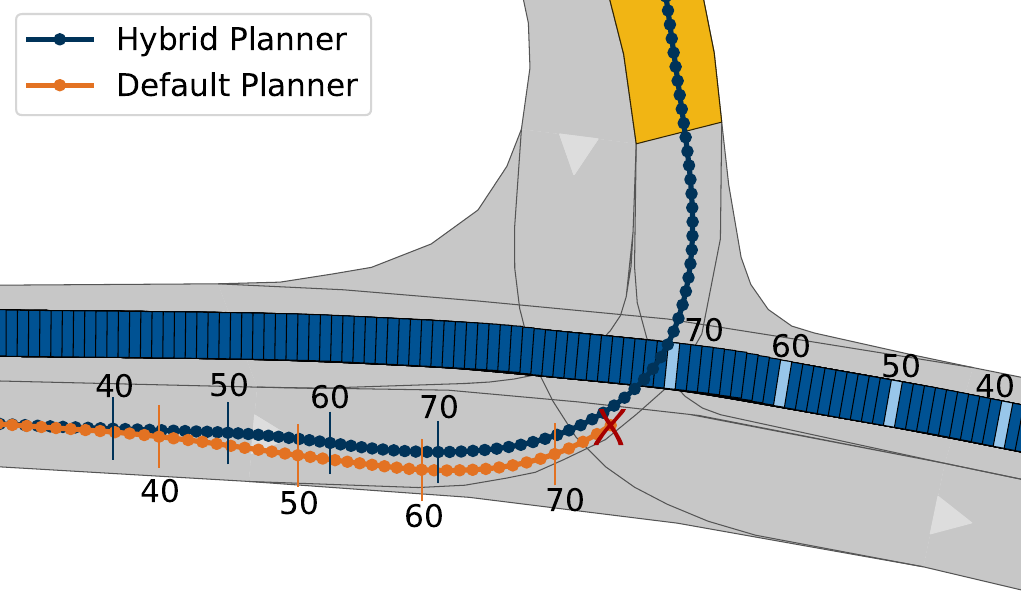}
  \caption{Turning left with HP (blue) and DP (orange) with oncoming traffic. Illustrated with time progression by timestamps. The collision of the DP is marked with a red cross.}
  \label{fig:pathdeviation}
\end{figure}

The HP makes it possible to adapt the driving behavior of the analytic trajectory planning algorithm during runtime. In the following analysis, we show the differences in driving behavior between the proposed model and the standalone analytical trajectory planner.
\cref{fig:pathdeviation} illustrates the HP compared to the DP in the same scenario in blue and orange, respectively.

Qualitatively, a strong adaptation of the driving behavior due to the oncoming vehicle can be determined. 
The center positions of the ego vehicle are shown according to the timestamp points. As the blue trajectory indicates, our methodology demonstrates improved adherence to the designated reference path, complemented by earlier braking initiation. In contrast, the DP drives with a greater deviation from the reference path but quickly approaches the oncoming vehicle. 
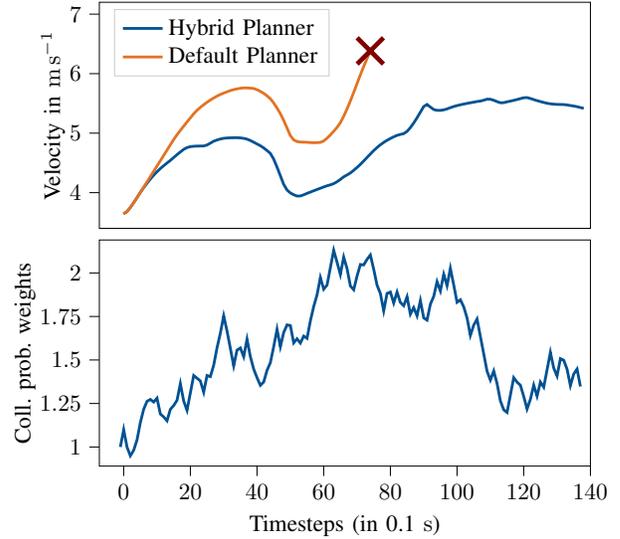
\begin{figure}[ht]
  \centering
  \scalebox{0.88}{
\begin{tikzpicture}

\definecolor{tumblue}{RGB}{0,82,147}
\definecolor{darkgray176}{RGB}{176,176,176}
\definecolor{lightgray204}{RGB}{204,204,204}
\definecolor{midnightblue05189}{RGB}{0,51,89}
\definecolor{orange}{RGB}{227,114,34}
\definecolor{red}{RGB}{128,0,0}

\begin{groupplot}[group style={group size=1 by 2, vertical sep=5pt}]
\nextgroupplot[
height=5cm,
width=9cm,
legend cell align={left},
legend style={fill opacity=0.8, at={(0.03,0.97)}, anchor=north west, draw opacity=1, text opacity=1, draw=lightgray204},
tick align=outside,
tick pos=left,
xtick=\empty,
xticklabels={},
xmin=-7.25, xmax=140.0,
y grid style={darkgray176},
ylabel style={yshift=-0.5em},
ylabel={Velocity in \SI{}{\meter\per\second}},
ymin=3.4, ymax=7.2,
ytick style={color=black}
]

\addplot [red, mark=x, mark size=8, mark options={solid,fill=black, 
        line width=2pt}, only marks, forget plot]
table {%
74 6.38244810530111
};
\addplot [very thick, tumblue]
table {%
0 3.64381350544534
1 3.67951709450729
2 3.74682880918198
3 3.8269882726638
4 3.91237784647222
5 3.99988235509629
6 4.0803928859649
7 4.15425404345871
8 4.22536041180688
9 4.28995243519146
10 4.34920722638841
11 4.40234912980623
12 4.45096267642271
13 4.49654011013771
14 4.54030618411821
15 4.58825373694648
16 4.63648924683358
17 4.68191988536111
18 4.72562788487455
19 4.75679742513437
20 4.77260406411861
21 4.77960122066579
22 4.78275749233863
23 4.78397942436177
24 4.78424684533181
25 4.80019904063764
26 4.83084486223976
27 4.86393024122496
28 4.88424626906487
29 4.90146342035841
30 4.91337776039339
31 4.91891854799773
32 4.92160283017828
33 4.92301781608275
34 4.92386602976724
35 4.92042787353169
36 4.91312813625651
37 4.90004991823797
38 4.87718054208731
39 4.84891243514773
40 4.81900173037565
41 4.7901417173915
42 4.75339319847347
43 4.70910395756263
44 4.66198074939448
45 4.56688890581843
46 4.43349974690153
47 4.28874410566113
48 4.14287707859767
49 4.04694981264735
50 3.99644424237157
51 3.96567794062471
52 3.94352926359507
53 3.94384961277662
54 3.96323307099795
55 3.98748960851193
56 4.00866950438916
57 4.03119115415572
58 4.05264038377881
59 4.07349504341602
60 4.09505445252641
61 4.11497204332579
62 4.1305127296617
63 4.1511886301632
64 4.18224827875768
65 4.22119139769037
66 4.26511386127313
67 4.29639731386023
68 4.33018806108036
69 4.37405620782076
70 4.42237771216236
71 4.47368740405077
72 4.5311025033009
73 4.59020317945543
74 4.64821830429381
75 4.70600291188096
76 4.7574135354531
77 4.80383252004905
78 4.84389909728736
79 4.87876644034934
80 4.90957003661757
81 4.93781868711041
82 4.96226740963997
83 4.97847706993145
84 4.99176137449721
85 5.02888683474824
86 5.09188989205729
87 5.16958325864753
88 5.25720043830434
89 5.35134353537828
90 5.44727094672057
91 5.48255437726727
92 5.43882989930982
93 5.39787349175948
94 5.38339688003192
95 5.38314609600011
96 5.39137103433942
97 5.40707404769143
98 5.42826836075178
99 5.44754779359459
100 5.46317646203215
101 5.47665833764298
102 5.48901193651241
103 5.50023160633957
104 5.51073787473696
105 5.52061089877824
106 5.5320478575722
107 5.53897468234632
108 5.55361718577565
109 5.57094514564372
110 5.57007170534644
111 5.55324706132621
112 5.52995133615504
113 5.51104850371815
114 5.51250596036948
115 5.52684459071625
116 5.53834204201452
117 5.55088994015804
118 5.56523698704039
119 5.58037121272857
120 5.59479243277869
121 5.59798120987935
122 5.58298847815446
123 5.56421576471125
124 5.54694214127337
125 5.53120483701515
126 5.5160971148744
127 5.5026434828989
128 5.49322878468105
129 5.48726021122906
130 5.48334080327834
131 5.48212527732906
132 5.48057497618861
133 5.47447503777674
134 5.46460721772068
135 5.455646965022
136 5.44454852434391
137 5.43079612340722
138 5.41653697595362

};
\addlegendentry{Hybrid Planner}
\addplot [very thick, orange]
table {%
0 3.64381350544534
1 3.67951709450729
2 3.74682880918198
3 3.8269882726638
4 3.91237784647222
5 3.99988235509631
6 4.08825630512856
7 4.17699735007766
8 4.26590044631059
9 4.35488047990481
10 4.44390060428709
11 4.53293445795883
12 4.62192550412563
13 4.7108725038994
14 4.79981162644025
15 4.88875154567043
16 4.96983054363009
17 5.04389245315789
18 5.11503387482511
19 5.184933745059
20 5.25412903940932
21 5.32287787671624
22 5.38237010453716
23 5.43178854217841
24 5.476040271219
25 5.51809203963183
26 5.55562783285154
27 5.58956739481799
28 5.62205854809683
29 5.6495399034603
30 5.67271902861197
31 5.69414320629256
32 5.71500592632882
33 5.73157290991981
34 5.74419053732497
35 5.75443975120776
36 5.76045177247387
37 5.76064604161341
38 5.75662832927638
39 5.75080905306619
40 5.73099533920995
41 5.70323882984332
42 5.67069506923285
43 5.62858466560013
44 5.580058815322
45 5.51007363165277
46 5.42345714683501
47 5.33138382801295
48 5.21814895968034
49 5.09089177503363
50 4.96149664584327
51 4.88755999161904
52 4.85985825250483
53 4.85044596169735
54 4.84935317443901
55 4.84715389054693
56 4.84178716095291
57 4.83917846708471
58 4.84069797300614
59 4.84394298848024
60 4.86850337025446
61 4.91356438445905
62 4.96891040558751
63 5.02986352663973
64 5.09518582886973
65 5.17375635299323
66 5.2813891105352
67 5.40368832103893
68 5.53809144218103
69 5.68192338220272
70 5.82946353264316
71 5.97667633166474
72 6.11822623691705
73 6.24989443619497
74 6.38244810530111
};
\addlegendentry{Default Planner}

\nextgroupplot[
height=5cm,
width=9cm,
tick align=outside,
tick pos=left,
x grid style={darkgray176},
xlabel={Timesteps (in 0.1 \SI{}{\second})},
xmin=-7.3, xmax=140.0,
xtick style={color=black},
y grid style={darkgray176},
ylabel={Coll. prob. weights},
ymin=0.890377440042794, ymax=2.18943319942802,
ylabel style={yshift=-0.5em},
ytick={1, 1.25, 1.5, 1.75, 2},
ytick style={color=black}
]
\addplot [very thick, tumblue]
table {
-1 1
0 1.1
1 1
2 0.949425429105759
3 0.980923888087273
4 1.04070931375027
5 1.14070931375027
6 1.21686863601208
7 1.26259939670563
8 1.27346586883068
9 1.25828637927771
10 1.28076813817024
11 1.18945855498314
12 1.17295057028532
13 1.15071919560432
14 1.21648474931717
15 1.23834720849991
16 1.26839609146118
17 1.3642246723175
18 1.2642246723175
19 1.21094589829445
20 1.31094589829445
21 1.41094589829445
22 1.39525554180145
23 1.37967563867569
24 1.31520367860794
25 1.41088970303535
26 1.404363489151
27 1.46982272267342
28 1.56982272267342
29 1.65363602638245
30 1.75363602638245
31 1.66660551428795
32 1.56660551428795
33 1.46660551428795
34 1.55672416090965
35 1.56898399442434
36 1.51912890225649
37 1.61912890225649
38 1.51912890225649
39 1.44746076017618
40 1.39831209927797
41 1.35440407544374
42 1.37355060577393
43 1.44043787717819
44 1.48449839651585
45 1.57630277574062
46 1.67630277574062
47 1.57630277574062
48 1.65774956047535
49 1.70158165693283
50 1.69807115569711
51 1.59807115569711
52 1.62185809239745
53 1.59706237986684
54 1.63713405802846
55 1.62416213527322
56 1.72416213527322
57 1.80704418793321
58 1.87388595119119
59 1.9736657358706
60 1.90660386458039
61 1.93038521036506
62 2.03038521036506
63 2.13038521036506
64 2.0617314003408
65 1.99130173549056
66 2.09130173549056
67 2.02548238858581
68 1.92548238858581
69 1.90334574654698
70 1.98194356635213
71 2.04850674942136
72 2.04609254784882
73 2.07803529389203
74 2.10335939116776
75 2.02121353335679
76 1.92972082681954
77 1.88229562230408
78 1.78229562230408
79 1.88229562230408
80 1.89038772173226
81 1.82742977924645
82 1.8903618235141
83 1.8307383377105
84 1.81159729249775
85 1.86294229514897
86 1.76294229514897
87 1.80099685676396
88 1.75378406532109
89 1.8413571882993
90 1.7413571882993
91 1.72892464287579
92 1.82075638063252
93 1.86530579216778
94 1.94980745203793
95 1.89666591770947
96 1.99188626892865
97 1.92681269533932
98 2.02681269533932
99 1.93134157545865
100 1.83134157545865
101 1.84582930095494
102 1.80523258931935
103 1.74146451242268
104 1.64146451242268
105 1.69985318668187
106 1.73523012883961
107 1.63523012883961
108 1.53523012883961
109 1.43523012883961
110 1.3846126306802
111 1.43674132116139
112 1.36499003656209
113 1.26499003656209
114 1.21400032527745
115 1.19855081625283
116 1.29855081625283
117 1.39855081625283
118 1.37258396930993
119 1.35528593845665
120 1.29059693403542
121 1.2182977873832
122 1.27387479133904
123 1.35565616078675
124 1.29236051626503
125 1.37336345501244
126 1.34356009848416
127 1.44356009848416
128 1.54356009848416
129 1.45359002836049
130 1.40781683214009
131 1.50781683214009
132 1.49870206750929
133 1.44685559310019
134 1.35278919376433
135 1.41959382928908
136 1.44782028235495
137 1.34782028235495

};
\end{groupplot}

\end{tikzpicture}
  }
  \vspace{-0.5em}
  \caption{Velocity profile of the HP and the DP and the relative changes in the collision probability weights due to the actions of the hybrid planner. The collision of the DP is marked with a red cross.}
  \label{fig:actionprediction}
\end{figure}

This accelerated approach results in an unintended violation of the vehicle's safety limits at the \nth{74} timestep, which leads to a collision with the oncoming vehicle. The scenario can be completed by carefully changing the manually set parameters of the DP. However, the results show that our HP can avoid manually tuning the parameters. \cref{fig:actionprediction} shows the velocity of the DP and HP and the HP agent's actions to adjust the planner's collision probability weights during the same scenario.
The velocity of the HP is significantly reduced compared to the DP so that no collision can occur in this situation. This is achieved by successively increasing the weights for the collision probability cost term through the agent's actions. The RL model can even partially compensate for conceptual errors in the cost function, which can be derived from the strong acceleration of the DP in this situation.

\begin{figure}[ht]
  \centering
  \scalebox{0.88}{
\begin{tikzpicture}[font=\small]

\definecolor{tumblue}{RGB}{0,82,147}
\definecolor{darkgray176}{RGB}{176,176,176}
\definecolor{lightgray204}{RGB}{204,204,204}
\definecolor{midnightblue05189}{RGB}{0,51,89}
\definecolor{orange}{RGB}{227,114,34}
\definecolor{red}{RGB}{128,0,0}

\begin{groupplot}[group style={group size=1 by 2}]
\nextgroupplot[
legend cell align={left},
legend style={fill opacity=0.8, at={(0.03,0.97)}, anchor=north west, draw opacity=1, text opacity=1, draw=lightgray204},
tick align=outside,
tick pos=left,
x grid style={darkgray176},
xlabel={Timesteps (in 0.1s)},
xmin=33.0, xmax=77.0,
xtick style={color=black},
y grid style={darkgray176},
ylabel={Ego risk},
ymin=-6.2999899890815e-05, ymax=0.00132299789770712,
ytick style={color=black}
]
\addplot [very thick, tumblue]
table {%
0 0
1 0
2 0
3 0
4 0
5 0
6 0
7 0
8 0
9 0
10 0
11 0
12 0
13 0
14 0
15 0
16 0
17 0
18 0
19 0
20 0
21 0
22 0
23 0
24 0
25 0
26 0
27 0
28 0
29 0
30 0
31 0
32 0
33 0
34 0
35 0
36 0
37 0
38 0
39 0
40 0
41 0
42 0
43 0
44 0
45 0
46 0
47 0
48 0
49 0
50 0
51 0
52 0.0005526209015964
53 0.0007220216035315
54 0
55 0
56 0
57 0
58 0
59 0
60 8.17985019777362e-09
61 0
62 1.15659518389585e-06
63 1.10707049673139e-05
64 3.37873523957401e-05
65 5.76539810766788e-06
66 3.98165774975843e-09
67 5.38723532860219e-11
68 2.78112543751157e-09
69 2.66751307829875e-13
70 1.70261255421672e-14
71 2.30037510214738e-10
72 1.06788734922387e-10
73 2.14285338936805e-13
74 8.70687181516675e-16
75 4.21462500464327e-17
76 3.25446657314891e-17
77 1.07110309492493e-16
78 5.76829134948168e-17
79 1.18515603140787e-15
80 4.54519553660825e-14
81 9.12046490356542e-13
82 4.09337901193169e-12
83 1.47707213848866e-11
84 3.34139058695598e-12
85 2.84331920383468e-14
86 9.59870975971294e-16
87 0
88 0
89 0
90 0
91 0
92 0
93 0
94 0
95 0
96 0
97 0
98 0
99 0
100 0
101 0
102 0
103 1.2258371772547e-08
104 8.57535057773537e-09
105 6.0360676092935e-12
106 3.35258540768871e-12
107 1.05838148809247e-11
108 3.73548067969868e-10
109 1.32423686798581e-11
110 9.8023999789354e-17
111 3.93353431495768e-17
112 3.75218964147656e-13
113 1.34733649074942e-14
114 4.02652938639593e-10
115 6.29235284632002e-09
116 6.93366757670138e-12
117 6.41365010553633e-11
118 2.45060364968142e-11
119 4.5794780911574e-12
120 1.06445118971497e-11
121 4.53089797094408e-12
122 5.3416393395207e-10
123 2.1310290113205e-11
124 9.24991857859024e-14
125 9.3937123303395e-16
126 2.04319800141855e-16
127 1.83019263740865e-15
128 9.03677158716439e-13
129 4.81952733356129e-16
130 6.20108565750523e-20
131 6.14211352722883e-20
132 5.55091574222436e-20
133 2.69503531711578e-20
134 6.17894294603099e-21
135 1.46224251468367e-23
136 2.84967472511167e-22
137 1.32288337565111e-20
138 6.18187769620221e-20
139 7.89829261024657e-20
140 1.25797341694449e-20
141 9.26366618367741e-22
142 7.57488663656775e-19
143 0
144 0
145 0
};
\addlegendentry{Hybrid Planner}
\addplot [very thick, orange]
table {%
0 0
1 0
2 0
3 0
4 0
5 0
6 0
7 0
8 0
9 0
10 0
11 0
12 0
13 0
14 0
15 0
16 0
17 0
18 0
19 0
20 0
21 0
22 0
23 0
24 0
25 0
26 0
27 0
28 0
29 0
30 0
31 0
32 0
33 0
34 0
35 0
36 0
37 7.5753552427312e-05
38 7.65552225895078e-05
39 0
40 0
41 0.0001026440653919
42 2.75509869746925e-05
43 6.02421674900451e-05
44 0
45 0.0004156956516649
46 0
47 0.0004105618340036
48 0
49 0.0005741976875927
50 0.0010024468449867
51 0.0005287573547904
52 0.0004296970099371
53 0.000604347141013
54 0.0001434864213536
55 0.000144097487451
56 0.0001054416262848
57 1.0392457219907e-05
58 7.24745157387747e-06
59 0.0012599979978163
60 0.0006000444651373
61 0.0009669353547746
62 0.0003319491064676
63 0.0003185363446118
64 0.0002467034109797
65 0.0001909873459936
66 2.75154907433785e-05
67 5.83500810745068e-05
68 6.29777469754213e-05
69 3.69061383862105e-05
70 2.95003667229026e-05
71 2.88449917229253e-05
72 0.000131973782946
73 0.0002093041101024
74 0.0002330472402173
};
\addlegendentry{Default Planner}

\addplot [red, mark=x, mark size=6, mark options={solid,fill=black, 
        line width=2pt}, only marks, forget plot]
table {%
74 0.0002330472402173
};

\end{groupplot}

\end{tikzpicture}
  }
  \caption{Ego vehicle risk distribution of the HP and DP algorithms~\cite{geisslingernature,GeißlerMaxRisk}. Risk is plotted over a specific time interval, with each step representing \SI{0.1}{\second}. The collision of the DP is marked with a red cross.}
  \label{fig:ego_risk_qualitative_analysis}
\end{figure}
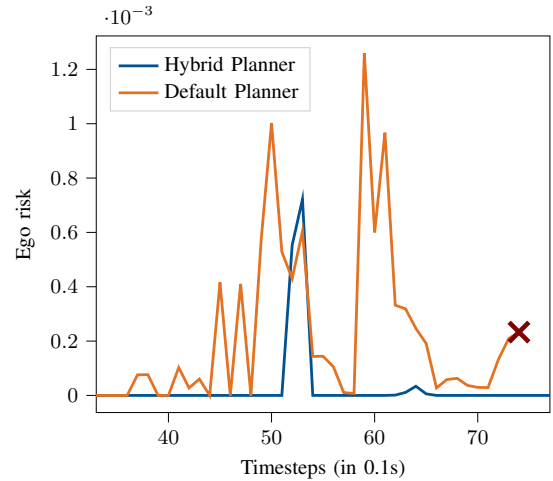

The active ego risk reduction during a scenario can be illustrated in \cref{fig:ego_risk_qualitative_analysis}. 
It can be seen that the sum at risk is significantly lower in our model. The theoretically calculated risk does not necessarily reflect the occurrence of a collision. 
However, the collisions are avoided by the model, and the risk, which is calculated with the potential harm, is minimized. Incorrect predictions of the objects cause the actions that lead to a collision of the DP. The results show that these can be compensated through the model.

\subsection{Scenario Performance Evaluation}
\begin{figure}[ht]
  \centering
  \includegraphics[width=0.485\textwidth]{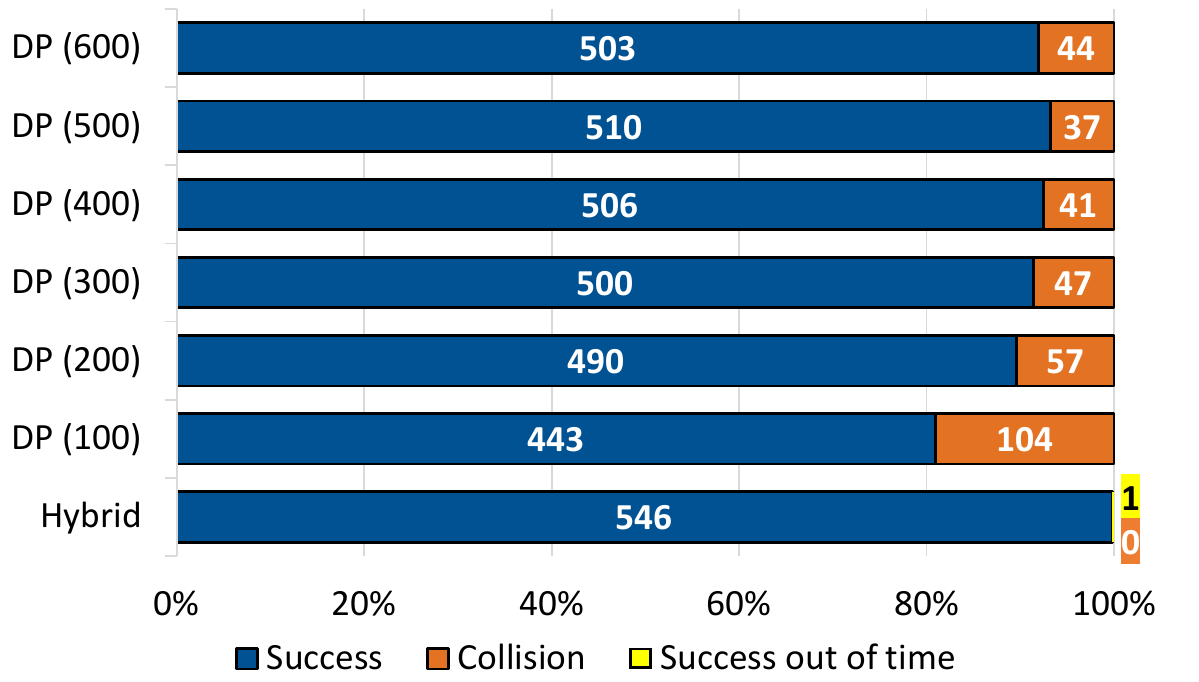}
  \vspace{-2.0em}
  \caption{Comparison of the DP and the HP with \SI{547}{} scenarios. The DP is executed with different collision probability costs to investigate different setups. Orange indicates the number of scenarios with collisions.}
  \label{fig:success_rate}
  \vspace{-0.5em}
\end{figure}

The next study evaluates the success rates of our method. Different cost parameters for collision probability are applied in the DP to ensure parameterization accuracy. The results are compared with the HP, as shown in \cref{fig:success_rate}.
It can be noted that the DP has a high success rate but has collisions in every configuration. The appropriate setting of collision probability costs is crucial to balance the algorithm. Costs set too low may lead to collisions from excessively aggressive driving. Conversely, excessively high costs might result in rear-end collisions due to overly cautious behavior. The DP lacks sufficient flexibility and requires further features for optimal performance. The trained HP performs exceptionally, with no collisions observed, even in previously unseen test scenarios.
Differences in driving behavior can be obtained from \cref{tab:measurements_table}.
\begin{table}[ht]
\centering
\caption{Driving characteristic measurements between the default and hybrid planner over various scenarios.}
\renewcommand{\arraystretch}{1.2}
    \begin{tabular}{|c|c|c|c|c|c|}
    \hline
    & Measurements & Max & Average & Median & $\sigma$ \\ \hline\hline
    \multirow{3}{*}{\rotatebox{90}{Default}} & d-position in \SI{}{\meter} & \SI{0.251}{} & \SI{-0.102}{} & \SI{-0.0188}{} & \SI{0.22}{} \\
    & Velocity in \SI{}{\meter\per\second} & \SI{9.784}{} & \SI{6.786}{} & \SI{6.77}{} & \SI{0.986}{} \\
    & Cost unweighted & \SI{578.87}{} & \SI{17.895}{} & \SI{3.07}{} & \textbf{34.07} \\ \hline
    \multirow{3}{*}{\rotatebox{90}{Hybrid}} & d-position in \SI{}{\meter} & \textbf{0.726} & \SI{-0.11}{} & \SI{-0.0126}{} & \SI{0.224}{} \\
    & Velocity in \SI{}{\meter\per\second} & \textbf{8.179} & \SI{5.98}{} & \SI{6.132}{} & \SI{0.9897}{} \\
    & Cost unweighted & \SI{2024}{} & \SI{20.07}{} & \SI{1.11}{} & \textbf{78.72} \\ 
    \hline
    \end{tabular}
\label{tab:measurements_table}
\end{table}
The HP exhibits enhanced adaptability regarding the maximum permissible deviation from the reference path. In addition, the maximum and average velocity is reduced to improve the turning maneuver for the T-junction scenarios. Furthermore, the costs associated with the optimal trajectory in the HP show a more substantial deviation. This increased deviation is feasible due to the application of variable weights, offering a more nuanced approach to trajectory optimization.

\subsection{Execution Time Evaluation}
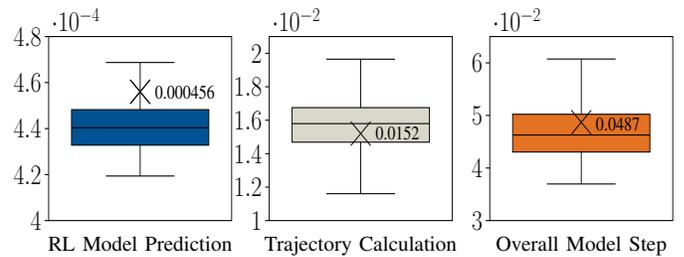
\begin{figure}[ht]
  \centering
  \setlength{\abovecaptionskip}{-15pt}

\begin{tikzpicture}[xscale=0.355, yscale=0.43]

\definecolor{chocolate22711434}{RGB}{227,114,34}
\definecolor{darkgray176}{RGB}{176,176,176}
\definecolor{lightgray218215203}{RGB}{218,215,203}
\definecolor{teal082147}{RGB}{0,82,147}

\begin{groupplot}[group style={group size=3 by 1, 
horizontal sep=40}]
\nextgroupplot[
tick align=outside,
tick pos=left,
xmin=0.6, xmax=1.4,
y grid style={darkgray176},
ymin=0.00040, ymax=0.00048,
yticklabel style={font=\fontsize{27}{12}\selectfont},
ytick style={color=black},
xtick=\empty
]
\path [draw=black, fill=teal082147]
(axis cs:0.7,0.000432789325714111)
--(axis cs:1.3,0.000432789325714111)
--(axis cs:1.3,0.000448226928710938)
--(axis cs:0.7,0.000448226928710938)
--(axis cs:0.7,0.000432789325714111)
--cycle;
\addplot [thick, black]
table {%
1 0.000432789325714111
1 0.000419378280639648
};
\addplot [thick, black]
table {%
1 0.000448226928710938
1 0.000468730926513672
};
\addplot [thick, black]
table {%
0.85 0.000419378280639648
1.15 0.000419378280639648
};
\addplot [thick, black]
table {%
0.85 0.000468730926513672
1.15 0.000468730926513672
};
\addplot [thick, black]
table {%
0.7 0.000440359115600586
1.3 0.000440359115600586
};
\addplot [black, mark=x, mark size=15, mark options={solid,fill=black}, only marks, forget plot]
table {%
1 0.000456
};
\addplot [black, mark=x, mark size=15, mark options={solid,fill=black}, only marks, forget plot]
table {%
1 0.000456
};
\node at (axis cs:1.05,0.000456) [right] {\LARGE 0.000456};

\nextgroupplot[
tick align=outside,
tick pos=left,
xmin=0.6, xmax=1.4,
y grid style={darkgray176},
yticklabel style={font=\fontsize{27}{12}\selectfont},
ymin=0.01, ymax=0.021,
ytick style={color=black},
xtick=\empty
]
\path [draw=black, fill=lightgray218215203]
(axis cs:0.7,0.0146880745887756)
--(axis cs:1.3,0.0146880745887756)
--(axis cs:1.3,0.0167514085769653)
--(axis cs:0.7,0.0167514085769653)
--(axis cs:0.7,0.0146880745887756)
--cycle;
\addplot [thick, black]
table {%
1 0.0146880745887756
1 0.0116052627563477
};
\addplot [thick, black]
table {%
1 0.0167514085769653
1 0.0196495056152344
};
\addplot [thick, black]
table {%
0.85 0.0116052627563477
1.15 0.0116052627563477
};
\addplot [thick, black]
table {%
0.85 0.0196495056152344
1.15 0.0196495056152344
};
\addplot [thick, black]
table {%
0.7 0.0157963037490845
1.3 0.0157963037490845
};
\addplot [black, mark=x, mark size=15, mark options={solid,fill=black}, only marks, forget plot]
table {%
1 0.0152
};
\node at (axis cs:1.05,0.01525) [right] {\LARGE 0.0152};

\nextgroupplot[
tick align=outside,
tick pos=left,
xmin=0.6, xmax=1.4,
y grid style={darkgray176},
yticklabel style={font=\fontsize{27}{12}\selectfont},
ymin=0.03, ymax=0.065,
ytick style={color=black},
xtick=\empty
]
\path [draw=black, fill=chocolate22711434]
(axis cs:0.7,0.0430405735969543)
--(axis cs:1.3,0.0430405735969543)
--(axis cs:1.3,0.0502225160598755)
--(axis cs:0.7,0.0502225160598755)
--(axis cs:0.7,0.0430405735969543)
--cycle;
\addplot [thick, black]
table {%
1 0.0430405735969543
1 0.0369627475738525
};
\addplot [thick, black]
table {%
1 0.0502225160598755
1 0.0607335567474365
};
\addplot [thick, black]
table {%
0.85 0.0369627475738525
1.15 0.0369627475738525
};
\addplot [thick, black]
table {%
0.85 0.0607335567474365
1.15 0.0607335567474365
};
\addplot [thick, black]
table {%
0.7 0.0462843179702759
1.3 0.0462843179702759
};
\addplot [black, mark=x, mark size=15, mark options={solid,fill=black}, only marks, forget plot]
table {%
1 0.0487
};
\addplot [black, mark=x, mark size=15, mark options={solid,fill=black}, only marks, forget plot]
table {%
1 0.0152
};
\node at (axis cs:1.05,0.0483) [right] {\LARGE 0.0487};

\end{groupplot}

\node [below] at ([yshift=-2mm]group c1r1.south) {\fontsize{8}{12}\selectfont RL Model Prediction};
\node [below] at ([yshift=-2mm]group c2r1.south) {\fontsize{8}{12}\selectfont Trajectory Calculation};
\node [below] at ([yshift=-2mm]group c3r1.south) {\fontsize{8}{12}\selectfont Overall Model Step};

\end{tikzpicture}
  \caption{Execution time per iteration of the RL model prediction, the trajectory bundle calculations, and the overall model in \SI{}{\second}.}
  \label{fig:boxplot_runtime}
  \vspace{-0.5em}
\end{figure}

\cref{fig:boxplot_runtime} illustrates the execution time per iteration in \SI{}{\second} for three critical components within the RL framework in the box-plot format: the RL model prediction execution, the sampling step of the DP, and the overall model execution. 
The calculation time is determined based on ten different scenarios. The average for the execution of agent prediction is approximately \SI{0.44}{\milli\second}. This step only includes the execution of the neural network and not the update of the environment model. An average of around \SI{15.8}{\milli\second} is required for the generation, validity check, and cost calculation of around \SI{800}{} trajectories per timestep. Increasing the number of trajectories in the analytical planning step has little impact on the computation time because the parallelized process is stable due to the C++ package extension. Running the entire model requires an average execution time of about \SI{46}{\milli\second} per iteration. 

\section{Discussion}
\label{sec:discussion}
The results show that the hybrid approach is effective and can significantly improve the analytical model with low execution times. In contrast to other pure RL models, the training process is fast and the model has a high success rate~\cite{Wang2021CR,shalevshwartz2016safe,Dong2023}. The generalisability increases significantly. Although the purely analytical model performs comparatively well in certain situations, performance can vary from situation to situation. 
In addition, with the right setup, the proposed model can compensate for the errors of other models, such as the prediction algorithm.
However, substantial alterations to the algorithm necessitate a partial retraining of the agent model. The design of the method can also be adapted and enhanced. Thus, the limits from \cref{eq:cost_weights} are frequently exploited, which indicates that the model can be improved. In addition, the choice of reward values and scenarios must be carefully considered, which can be time-consuming.
Overall, our concept demonstrates the effective use of the synergies offered by the hybrid planner and extends the currently available methods through greater complexity and applicability in edge-case scenarios~\cite{Klimke2023,Yu2023,Jafari2023}.

\section{Conclusion \& Outlook}
\label{sec:conclusion}
This paper introduces a hybrid motion planner approach for trajectory planning to enhance vehicle driving behavior under changing conditions. Addressing the low generalizability of traditional analytical trajectory planners, our method combines a sampling-based planner with an RL agent. This agent dynamically adjusts the cost weights in the analytical algorithm, boosting its adaptability.
Our approach utilizes an observation space, including the environmental, semantic map, and obstacle data, which are crucial for the hybrid agent's learning about vehicle dynamics. The results show a significant improvement in the agent's success rate and a reduction in risk while maintaining high-performance execution times for real-world applications.
Nevertheless, additional features could improve driving behavior and model performance through more extensive investigations. Future work could optimize the sampling parameters of the analytical planner using RL and thus investigate the algorithm's applicability in the real world. Including more comprehensive environmental data, e.g., through graph representations, could further increase the stability and efficiency of the system.


\end{document}